\documentclass{article}

\usepackage{makecell}
\usepackage{PRIMEarxiv}
\usepackage{tabularx}
\usepackage{amsmath}
\usepackage[utf8]{inputenc} 
\usepackage[T1]{fontenc}    
\usepackage{hyperref}       
\usepackage{url}            
\usepackage{booktabs}       
\usepackage{amsfonts}       
\usepackage{nicefrac}       
\usepackage{microtype}      
\usepackage{lipsum}
\usepackage{fancyhdr}       
\usepackage{graphicx}       
\graphicspath{{media/}}     
\usepackage{booktabs}
\usepackage{array}
\usepackage{longtable}
\title{GreenIQ: A Deep Search Platform for Comprehensive Carbon Market Analysis and Automated Report Generation }

\author{
  \makebox[3cm][c]{Oluwole Fagbohun} \\ 
  Carbonnote, \\ USA 
  \And
  \makebox[3cm][c]{Sai Yashwanth} \\ 
  Vuhosi Limited, \\ UK 
  \And
  \makebox[3cm][c]{Akinyemi Sadeeq Akintola} \\ 
  Universidade NOVA, \\ Portugal 
  \And
  \makebox[3cm][c]{Ifeoluwa Wurola} \\ 
  University of Hull, \\ UK 
  \AND
  \makebox[3cm][c]{Lanre Shittu} \\ 
  Vuhosi Limited, \\ UK 
  \And
  \makebox[3cm][c]{Aniema Inyang} \\ 
  Readrly Limited, \\ UK 
  \And
  \makebox[3cm][c]{Oluwatimilehin Odubola} \\ 
  ResilStudio Ltd, \\ UK 
  \And
  \makebox[3cm][c]{Udodirim Offia} \\ 
  Readrly Limited, \\ UK 
  \AND
  \makebox[3cm][c]{Said Olanrewaju} \\ 
  Cnergy Capital Ltd., \\ UK 
  \And
  \makebox[3cm][c]{Ogidan Toluwaleke} \\ 
  Vuhosi Limited, \\ UK 
  \And
  \makebox[3cm][c]{Ilemona Abutu} \\ 
  Carbonnote, \\ USA 
  \And
  \makebox[3cm][c]{Taiwo Akinbolaji} \\ 
  Silveredge Consults, \\ UK  
}

\begin{document}
\maketitle
\begin{abstract}
This study introduces GreenIQ, an AI-powered deep search platform designed to revolutionise carbon market intelligence through autonomous analysis and automated report generation. Carbon markets operate across diverse regulatory landscapes, generating vast amounts of heterogeneous data from policy documents, industry reports, academic literature, and real-time trading platforms. Traditional research approaches remain labour-intensive, slow, and difficult to scale. GreenIQ addresses these limitations through a multi-agent architecture powered by Large Language Models (LLMs), integrating five specialised AI agents: a Main Researcher Agent for intelligent information retrieval, a Report Writing Agent for structured synthesis, a Final Reviewer Agent for accuracy verification, a Data Visualisation Agent for enhanced interpretability, and a Translator Agent for multilingual adaptation. The system achieves seamless integration of structured and unstructured information with AI-driven citation verification, ensuring high transparency and reliability. GreenIQ delivers a 99.2\% reduction in processing time and a 99.7\% cost reduction compared to traditional research methodologies. A novel AI persona-based evaluation framework involving 16 domain-specific AI personas highlights its superior cross-jurisdictional analytical capabilities and regulatory insight generation. GreenIQ sets new standards in AI-driven research synthesis, policy analysis, and sustainability finance by streamlining carbon market research. It offers an efficient and scalable framework for environmental and financial intelligence, enabling more accurate, timely, and cost-effective decision-making in complex regulatory landscapes.
\end{abstract}


\vspace{1\baselineskip}
\hrule
\vspace{1\baselineskip}
\textbf{NOTE: Preprint, Submitted to ICMLT 2025, Helsinki, Finland}

\vspace{0.2\baselineskip}

Oluwole Fagbohun is with Carbonnote, Maryland, USA (e-mail: wole@carbonnote.ai).
Sai Yashwanth is with Vuhosi Limited, London, United Kingdom (e-mail: taddishetty34@gmail.com).
Akinyemi Sadeeq Akintola is with Universidade NOVA de Lisboa, Lisbon, Portugal (e-mail: sadeeq2@gmail.com).
Ifeoluwa Wurola is with the University of Hull, Hull, United Kingdom (e-mail: wuraolaifeoluwa@gmail.com).
Lanre Shittu is with Vuhosi Limited, London, United Kingdom (e-mail: hr.lanreshittu@gmail.com).
Aniema Inyang is with Readrly Limited, London, United Kingdom (e-mail: annywillow@gmail.com).
Oluwatimilehin Odubola is with ResilStudio Ltd, London, United Kingdom (e-mail: timi@resilstudio.com).
Udodirim Offia is with Readrly Limited, London, United Kingdom (e-mail: udodirim.offia@gmail.com).
Said Olanrewaju is with Cnergy Capital Limited, London, United Kingdom (e-mail: said.olar@cnergyfund.com).
Ogidan Toluwaleke is with Vuhosi Limited, London, United Kingdom (e-mail: gbemilekeogidan@gmail.com).
Ilemona Abutu is with Carbonnote, Maryland, USA (e-mail: abutu.mona@gmail.com).
Taiwo Akinbolaji is with Silveredge Consults, London, United Kingdom (e-mail: taiwoakinbolaji@gmail.com).

\vspace{0.2\baselineskip}
\hrule

\section{Introduction}
    The global carbon market has evolved significantly over the past two decades, transitioning from conceptual frameworks to complex, multi-jurisdictional trading systems with profound economic and environmental implications \cite{WorldBank2023,Newell2014}. This evolution has coincided with an increasingly urgent climate crisis, as evidenced by rising global temperatures and extreme weather events \cite{IPCC2023}, prompting governments worldwide to establish more ambitious regulatory frameworks and market-based mechanisms to reduce greenhouse gas emissions \cite{EUETS2023}. As these markets mature and expand, with the global carbon market value projected to reach $\$ 2.4$ trillion by 2027 \cite{Persefoni2024} the analytical challenges confronting stakeholders have grown exponentially.\\[0pt]
Carbon markets operate at the intersection of environmental science, economics, policy, and technology, generating heterogeneous data across disparate sources including regulatory documentation, academic literature, industry reports, and real-time trading platforms. For policymakers, investors, corporations, and researchers, the ability to synthesise and interpret this information efficiently has become a critical determinant of effective decision making \cite{Hepburn2020}. However, conventional approaches to carbon market analysis frequently prove inadequate and are characterised by labour-intensive manual research processes, significant time lags between market developments and analytical outputs, and inherent limitations in information processing capacity \cite{Fankhauser2010,Keohane2015}.\\[0pt]
The information asymmetry resulting from these inefficiencies has substantial consequences. Regulatory bodies risk implementing suboptimal policy frameworks without comprehensive market intelligence \cite{OECD2023}. Investors may miss strategic opportunities or misallocate capital because of incomplete market understanding \cite{Carney2021}. Corporations navigating compliance obligations or voluntary carbon commitments face uncertainty regarding market dynamics and regulatory trajectories \cite{TCFD2023}. Academic researchers may struggle to incorporate the latest market developments into their theoretical frameworks and empirical analyses \cite{Aldy2011}.\\[0pt]
Recent advancements in artificial intelligence, particularly in Large Language Models (LLMs), present promising opportunities to address these challenges \cite{Brown2020,Bommasani2022}. LLMs have demonstrated remarkable capabilities in natural language understanding, information synthesis, and content generation across various domains \cite{Zhao2023,Ouyang2022}. However, their application to specialised fields, such as carbon market analysis, has remained limited, with existing solutions typically focusing on narrow aspects of the analytical process rather than offering comprehensive end-to-end solutions \cite{Li2022,Bubeck2023}.\\
This study introduces GreenIQ, a novel deep search platform designed to transform carbon market intelligence through autonomous, comprehensive analysis, and report generation. GreenIQ leverages a sophisticated multiagent architecture powered by state-of-the-art LLMs, where each agent is engineered to perform specialised functions within an integrated analytical pipeline. The platform autonomously processes diverse data sources, including academic literature, regulatory documentation, industry reports, and real-time market data, generating detailed citation-rich reports tailored to specific stakeholder requirements.\\
The multi-agent framework underpinning GreenIQ comprises five distinct components: 
\begin{enumerate}
    \item A Main Researcher Agent responsible for comprehensive information gathering, trend identification, and source verification; 
    \item A Report Writing Agent that converts research findings into coherent, structured narratives; 
    \item A Final Reviewer Agent that ensures accuracy, consistency, and adherence to reporting standards; 
    \item A Data Visualization Agent that transforms complex data into accessible visualisations; and 
    \item A Translator Agent that enables multilingual report dissemination while preserving technical accuracy and contextual nuances.
    \end{enumerate}

Our contributions to the field are fourfold. First, we demonstrate the effective integration of advanced Large Language Models (LLMs) for specialised domain analysis, establishing a methodological framework for applying these technologies to complex, data-intensive research challenges. Second, we introduce a novel multi-agent architecture that enables autonomous, end-to-end research and reporting, significantly reducing the time and resources required for comprehensive carbon market analysis. Third, we provide empirical evidence of our approach's efficiency, accuracy, and cost-effectiveness through rigorous comparative evaluation against traditional manual research methodologies. Fourth, we introduce a novel AI persona-based evaluation framework, employing 16 domain-specific AI personas to assess analytical capabilities, citation quality, and regulatory insight generation in real-world reports written by humans. To ensure a comprehensive performance assessment, these AI personas were designed to mimic experts across various industries, enabling a consistent and scalable evaluation without the logistical constraints of assembling large expert panels. This innovative approach provides a robust and replicable methodology for evaluating AI-driven research frameworks from diverse stakeholder perspectives.\\
The remainder of this paper is organised as follows: Section 2 reviews relevant literature on carbon market analysis, automated research methodologies, and LLM applications in specialised domains. Section 3 details GreenIQ system architecture including the technical specifications and operational dynamics of each agent. Section 4 presents our evaluation methodology and results. Section 5 discusses the broader implications, limitations, and future research directions, followed by concluding remarks in Section 6.

\section{Related Work}
\subsection{Traditional Carbon Market Analysis}
Carbon market analysis has traditionally relied on manual research methodologies conducted by domain experts, who collect, synthesise, and interpret data from diverse sources. These processes typically involve consultancies, research institutions, and specialised market intelligence firms that produce periodic reports through labour-intensive workflows. Governmental bodies such as the World Bank and international organisations such as the International Carbon Action Partnership (ICAP) publish annual or biannual comprehensive assessments that serve as industry benchmarks \cite{ICAP2023}. While these reports offer valuable insights, they are characterised by significant time lags between market developments and published analyses, often becoming outdated shortly after release.\\[0pt]
Commercial platforms such as ICIS Carbon, Bloomberg New Energy Finance, and S\&P Global Platts provide more frequent market updates and analytical tools. However, these solutions generally focus on quantitative metrics and trading data rather than on comprehensive qualitative analyses of regulatory developments, academic research, and emerging trends. Additionally, McKinsey \cite{Blaufelder2021} notes that these platforms typically require substantial human intervention to contextualise data and generate actionable insights, limiting their scalability and responsiveness to rapidly evolving market conditions.\\[0pt]
Narassimhan et al. \cite{Narassimhan2018} highlight that existing analytical approaches struggle with the heterogeneity of carbon market data, which spans structured trading information, semi-structured regulatory documents, and unstructured textual sources such as news articles and research papers. This diversity presents significant challenges for traditional analytical methodologies, particularly when addressing the interconnected nature of global carbon markets across jurisdictions, sectors, and time horizons \cite{Doda2019}.

\subsection{AI in Environmental and Financial Analysis}
The application of artificial intelligence to environmental and financial analysis has expanded considerably in recent years. Machine learning techniques have demonstrated efficacy in various environmental domains, including climate prediction, emissions forecasting, and renewable energy optimisation \cite{Rolnick2019}. In financial contexts, AI systems routinely analyse market trends, assess risk factors, and generate trading signals \cite{Masood2024}, with increasing attention paid to climate-related financial considerations \cite{Bolton2020}. Rolnick et al. \cite{Rolnick2019} surveyed machine learning applications across climate change mitigation and adaptation strategies. In financial contexts, Lewis and Young \cite{Lewis2018} demonstrated the effectiveness of NLP techniques for analysing sentiment in regulatory filings and assessing their market impacts.\\[0pt]
The emergence of Large Language Models (LLMs) has significantly advanced research synthesis and knowledge integration capabilities. Recent studies have explored LLM applications in regulatory compliance analysis \cite{Hassani2024,Kothandapani2025}, and evidence synthesis for policy development \cite{Cao2024}. However, as noted by Bommasani et al. \cite{Bommasani2022}, these technologies face challenges when applied to highly specialised domains with technical terminology and complex conceptual frameworks. Carbon markets represent one such domain, requiring sophisticated understanding of environmental economics, regulatory frameworks, and market mechanisms \cite{Pahle2018}.

\subsection{Multi-Agent Systems for Research and Automated Information Retrieval Systems}
Multi-Agent Systems (MAS) have become pivotal in modelling complex interactions within environmental and financial domains. In environmental research, MAS facilitate the simulation of ecological systems and the assessment of climate policies. For instance, research has demonstrated the application of MAS in modelling climate change scenarios, providing insights into mitigation and adaptation strategies \cite{Weyns2024}. By simulating the interactions of various stakeholders, including governments, corporations, and consumers, MAS offer predictive capabilities that can support policy development and regulatory compliance. These simulations help identify the long-term impact of different climate policies, ensuring that decision-makers are equipped with robust, datadriven insights. In the financial sector, MAS have been utilised to predict global financial crises by simulating interactions among diverse market participants, thereby identifying systemic risks \cite{Patrick2024}. The ability of MAS to model complex financial ecosystems allows regulators and analysts to test market stability under various economic conditions. For example, by simulating speculative bubbles and liquidity constraints, MAS frameworks can anticipate vulnerabilities in financial systems and inform proactive regulatory interventions. These applications underscore MAS's capability to model complex systems, enhance decision-making, and provide scalable solutions across various sectors.

In the field of Information Retrieval (IR) systems, LLMs enhance IR by improving query understanding and document retrieval accuracy, allowing researchers to process large volumes of structured and unstructured data efficiently. Recent studies have explored the synergy between LLMs and IR, highlighting the potential to build autonomous agents that fully automate the information acquisition process \cite{Feng2023}. By leveraging retrievalaugmented generation (RAG), LLM-enhanced IR systems can dynamically retrieve relevant documents, synthesise knowledge, and generate context-aware responses, reducing the need for manual information curation. Additionally, LLMs have been employed to evaluate IR systems, demonstrating their capability to assess retrieval performance effectively \cite{Rahmani2023}. Traditional IR systems relied on precision-recall metrics, but LLM-based evaluation frameworks now enable context-aware relevance assessments, refining search accuracy. These advancements have transformed traditional IR systems, making them more adaptive and capable of handling complex, multi-faceted queries. As a result, automated IR systems have become essential for regulatory analysis, financial oversight, and climate policy research, enhancing research efficiency and effectiveness in high-stakes decision-making environments.

\subsection{Gap Analysis}
Despite the increasing sophistication of carbon market intelligence tools, existing research methodologies continue to exhibit critical limitations that hinder their effectiveness in a rapidly evolving regulatory landscape. Traditional expert-driven approaches, while historically valuable, remain constrained by laborious manual workflows, significant time delays, and inherent scalability issues. Analysts must painstakingly sift through disparate regulatory documents, financial reports, and scientific publications-an effort that not only consumes substantial resources but also results in insights that risk obsolescence by the time they are disseminated. This bottleneck is particularly problematic in the context of carbon markets, where policy changes, trading dynamics, and compliance frameworks evolve at an accelerated pace \cite{Borri2024}. Meanwhile, commercial research platforms, though offering incremental efficiency gains, often focus narrowly on structured data-emissions figures, price movements, or transactional records - neglecting the broader qualitative context that drives market behaviours \cite{ICAP2023}. Even when artificial intelligence is deployed within existing platforms, its role is largely limited to basic keyword extraction or sentiment analysis, failing to capture the nuanced interplay of financial, regulatory, and environmental variables that shape the carbon market ecosystem \cite{WJARR2024}. Moreover, voluntary carbon markets continue to face significant challenges regarding data reliability, uncertainty in carbon credit valuations, and the standardisation of offset mechanisms, further exacerbating inefficiencies in traditional carbon market intelligence methods \cite{MIT2023}.\\[0pt]
The recent emergence of Large Language Models (LLMs) and Multi-Agent Systems (MAS) has opened new possibilities for automating carbon market research, yet current AI-driven frameworks suffer from several unresolved challenges. Many LLM-based tools, while excelling at textual synthesis, struggle with fact verification and citation integrity, occasionally generating plausible yet misleading narratives-an issue that regulatory analysts and financial decision-makers cannot afford \cite{Oladeji2023}. Furthermore, existing AI-driven solutions lack the necessary modularity to adapt seamlessly to multi-jurisdictional carbon pricing mechanisms, which vary significantly in scope, governance, and economic impact \cite{Hassani2024}. In the absence of a robust multi-agent coordination framework, these systems often operate in silos, failing to integrate real-time trading data with evolving policy regulations in a cohesive manner \cite{Patrick2024}. GreenIQ directly addresses these gaps by implementing a deep search platform that not only automates the end-to-end research pipeline but also embeds citation verification, jurisdictional adaptability, and stakeholder-specific contextualisation into its design. By leveraging a five-agent architecture - each specialising in a distinct research function-GreenIQ ensures that carbon market intelligence is not only comprehensive and timely but also rigorously validated, democratising access to highquality, AI-driven sustainability analysis.

\section{GreenIQ Architecture and Methodology}
\subsection{System Overview}
\begin{figure}[h]
    \centering
\includegraphics[width=0.5\textwidth]{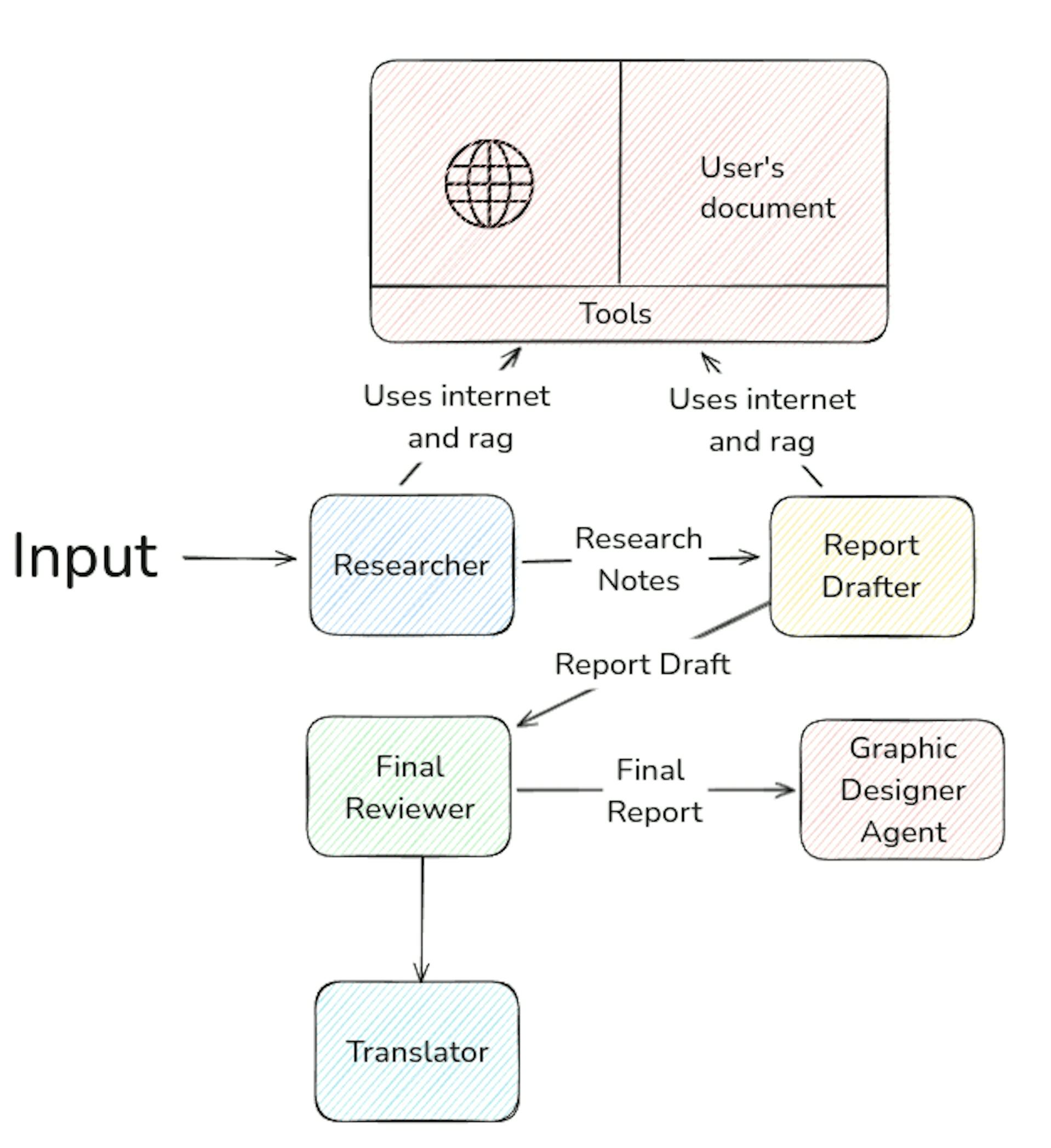}  
    \caption{High-level architecture diagram of the GreenIQ system, illustrating the modular design and data flow between the five autonomous agents.}
    \label{fig:greeniq_architecture}
\end{figure}

GreenIQ is an advanced deep search platform designed to automate carbon market intelligence and reporting through an integrated multi-agent system. The architecture follows a modular design as shown in figure \ref{fig:greeniq_architecture},\\
where each specialised agent contributes to a structured, end-to-end workflow from data collection to multilingual reporting.\\

A key feature of GreenIQ is its deep search functionality, which integrates data from diverse sources, including regulatory repositories, academic literature, industry reports, real-time market data, and user-provided proprietary datasets. The data ingestion pipeline is structured into multiple stages, beginning with source identification and validation, followed by structured extraction using tailored parsers for different document types. The semantic analysis layer plays a crucial role in refining retrieved data by leveraging Named Entity Recognition (NER), topic modelling, and relationship extraction techniques. Additionally, iterative query refinement ensures that the system continuously improves its accuracy and relevance by dynamically adjusting search parameters based on user needs. Users can also define specific date ranges to retrieve only time-sensitive market insights, improving the precision of regulatory and financial intelligence. This streamlined pipeline enables the rapid generation of structured, accurate, and multilingual reports that would traditionally require extensive manual research and expert review.

\subsection{Multi-Agent Framework}
GreenIQ employs a robust multi-agent architecture, where each agent is engineered with specialised tasks, ensuring efficient automation of the entire research and reporting process.

\subsubsection{Main Researcher Agent}
The Main Researcher Agent is responsible for comprehensive information gathering. It integrates APIs, web scraping tools, and retrieval-augmented generation (RAG) models to extract relevant data from authoritative sources such as regulatory bodies, academic journals, and financial institutions. The agent systematically analysis source credibility, ensuring all extracted data aligns with industry standards. The output from this agent is a structured research report, complete with citations and annotations.

\subsubsection{Report Agent}
The Report Agent synthesises findings from the research report, structuring them into an engaging and logically coherent narrative. By utilising content organisation strategies, it adapts report structures for different\\
stakeholders, such as policymakers, investors, and sustainability analysts. The agent dynamically adjusts language complexity, ensuring clarity for technical and non-technical audiences alike.

\subsubsection{Final Reviewer Agent}
To uphold quality control standards, the Final Reviewer Agent performs meticulous fact-checking, citation validation, and linguistic refinement. This agent ensures that every claim is backed by a verifiable source, citations comply with academic and industry norms, and the overall report adheres to professional writing standards. It also detects potential biases or inconsistencies, mitigating risks associated with automated content generation.

\subsubsection{Data Visualisation Agent}
The Data Visualisation Agent converts complex data insights into automated charts, graphs, and infographics. By leveraging Plotly and custom-built visualisation libraries, this agent enhances the interpretability of carbon market data. It dynamically selects the most appropriate visual formats-such as bar charts, heatmaps, and network graphs-based on the underlying dataset.

\subsubsection{Translator Agent}
To ensure accessibility across different markets, the Translator Agent adapts the report into multiple languages. Beyond direct translation, it employs contextual adaptation techniques to account for regional terminologies and regulatory nuances. This guarantees that translated versions maintain technical precision while being culturally and linguistically appropriate.

\subsection{Technical Implementation}
GreenIQ's technical infrastructure is powered by a combination of state-of-the-art LLMs, retrieval-enhanced search tools, and custom-built AI agents, designed for seamless interoperability. The Main Researcher Agent utilises GPT-4o as its core model, integrating an internet search tool for real-time data retrieval, a retrievalaugmented generation (RAG) mechanism for incorporating user-provided datasets, and a file writing tool to document findings systematically. The Report Drafter Agent and the Final Reviewer Agent are built on GPT-$40-m i n i$, specialising in content organisation, linguistic refinement, and consistency checks. The Data Visualisation Designer employs GPT-4o in conjunction with Plotly-based dynamic visualisation tools to create meaningful graphical representations of complex data. The Translator Agent, powered by GPT-4o-mini, is equipped with file writing and multilingual adaptation tools, ensuring reports are accurately localised for diverse audiences.\\
The system follows a linear processing pipeline, where each agent's output becomes the next agent's input. The workflow begins with the Main Researcher Agent, which compiles structured research reports by integrating information from regulatory and academic sources. The Report Drafter Agent then transforms these raw findings into a structured report, refining narratives to enhance coherence and clarity. The Final Reviewer Agent verifies the accuracy of the content, ensuring citation integrity and compliance with academic and industry standards. Once the textual content is validated, the Data Visualisation Designer generates graphical representations of key insights, improving the accessibility and impact of the report. Finally, the Translator Agent adapts the document for international stakeholders, maintaining linguistic precision and cultural relevance. Each agent operates autonomously within this tightly integrated workflow, ensuring efficiency, accuracy, and adaptability. Extensive testing has confirmed the optimal performance of these models, with seamless interoperability across all agents, allowing GreenIQ to function as a scalable and reliable deep search platform for carbon market intelligence.

\section{Evaluation and Results}
\subsection{Evaluation Methodology}
Our evaluation methodology was designed to rigorously assess GreenIQ performance across multiple dimensions relevant to carbon market analysis and reporting. We established a comprehensive framework that combines structured scoring criteria with a novel AI persona evaluation approach.

\subsubsection{Scoring Framework}
The evaluation of reports generated by GreenIQ (Report 1), and baseline human reports (Report 2) employed four key criteria, each scored on a 25 -point scale for a total possible score of 100 points per report:

\begin{enumerate}
  \item Source Coverage ( 25 points)
\end{enumerate}

\begin{itemize}
  \item Strengths: High scores indicate diverse source utilisation (regulatory repositories, academic literature, industry reports, market data), providing comprehensive topic coverage with both primary and secondary sources.
  \item Weaknesses: Low scores result from limited source variety, insufficient information depth, or failure to reference key regulatory or academic sources.
\end{itemize}

\begin{enumerate}
  \setcounter{enumi}{1}
  \item Data Accuracy ( 25 points)
\end{enumerate}

\begin{itemize}
  \item Strengths: High scores reflect current properly sourced data free from errors, with clear methodologies for collection and analysis.
  \item Weaknesses: Low scores indicate outdated information, improper sourcing, errors in data presentation, unsupported claims, or lack of statistical context.
\end{itemize}

\begin{enumerate}
  \setcounter{enumi}{2}
  \item Citation Quality (25 points)
\end{enumerate}

\begin{itemize}
  \item Strengths: High scores indicate citations from authoritative, reputable sources, including peerreviewed articles, government publications, and recognised industry reports.
  \item Weaknesses: Low scores reflect missing citations, reliance on unreliable sources, nonauthoritative references, or complete lack of citations.
\end{itemize}

\begin{enumerate}
  \setcounter{enumi}{3}
  \item Report Coherence ( 25 points)
\end{enumerate}

\begin{itemize}
  \item Strengths: High scores indicate well organised reports with a clear introduction, body, and conclusion featuring logical narrative flow and easy-to-follow argumentation.
  \item Weaknesses: Low scores result from poor organisation, unclear transitions, disconnected topics, or incoherent argumentation.
\end{itemize}

\subsubsection{AI Persona Evaluation Framework}
To evaluate GreenIQ's performance across diverse stakeholder perspectives comprehensively, we developed 16 AI personas designed to emulate domain experts from various industries. This novel evaluation framework ensures consistent, scalable, and reproducible assessment while mitigating the logistical challenges associated with assembling large human expert panels. The use of AI personas as evaluators is an emerging approach in research, with recent studies demonstrating their effectiveness in replicating human expert assessments across various domains \cite{Sun2024,Zhou2024}. Our methodology aligns with findings from prior research, where AI personas were successfully employed to mirror human reasoning and domain expertise in experimental replications \cite{Joseph2024}.\\[0pt]
Our selection of 16 distinct AI personas was strategically grounded in achieving comprehensive coverage across key expertise domains relevant to carbon markets, including regulatory frameworks, economic analyses, scientific research, sustainability policies, and industry-specific applications. Each AI persona was assigned to a specific expertise domain and powered by a corresponding foundation model optimised for domain-specific reasoning. The Climate Policy and Regulatory Experts group (powered by Claude 3 Opus Mini) included specialists in international climate agreements, jurisdiction-specific carbon regulatory frameworks, and policy implementation strategies. The Environmental and Economic Analysis group (powered by Claude 3.1 Mini) consisted of environmental economists analysing carbon pricing mechanisms, carbon market strategists focusing on offset structures, and climate risk modellers forecasting economic implications. The Scientific and Technical Experts group (powered by GPT-4.5) incorporated atmospheric scientists with climate modelling expertise, climate data scientists leveraging machine learning for predictive analytics, and environmental journalists specialising in scientific fact-checking. The Sustainability and Finance Specialists group (powered by GPT-4o) comprised ESG investment analysts assessing corporate sustainability, climate resilience strategists developing adaptation frameworks, and biodiversity impact specialists evaluating climate-induced ecosystem changes. Finally, the Specialised Perspectives group (powered by GPT-4o Mini) included renewable energy technology specialists, climate justice advocates examining socio-environmental impacts, and AI agent developers designing decision-support systems for climate governance. These AI personas were inspired by methodologies such as those employed in Writer-Defined AI Personas for On-Demand Feedback Generation, where diverse personas were constructed to simulate audience perspectives in writing feedback \cite{Sun2024}.

To mitigate model-specific biases, we deliberately employed multiple foundation models, a methodological approach supported by recent meta-analyses of LLM evaluation frameworks \cite{Zhou2024}. Each model family exhibits distinct performance characteristics across specialised knowledge domains, necessitating cross-model\\[0pt]
triangulation to enhance evaluation validity and robustness. By diversifying the model architectures (Claude and GPT variants), we minimised systematic errors arising from idiosyncratic knowledge representations or reasoning pathways within any single model family. Furthermore, each AI persona was provided with a predefined domain perspective but did not undergo additional training or fine-tuning beyond the base model architecture. This design decision aligns with findings from From Persona to Personalization: A Survey on Role-Playing Language Agents, which highlighted that AI personas retain their contextual accuracy without the need for extensive fine-tuning \cite{Joseph2024}. To ensure methodological rigour and unbiased assessment, we implemented a double-blind evaluation protocol, in which neither the AI personas nor human supervisors were aware of whether the reports originated from GreenIQ or baseline sources until after all evaluations were completed. Such an approach builds upon existing best practices in AI-assisted evaluations, as demonstrated in Using Large Language Models to Create AI Personas for Replication and Prediction of Media Effects, where AI personas successfully replicated human experimental assessments \cite{Zhou2024}. By incorporating these advances, GreenIQ's evaluation methodology represents a significant step towards scalable, AI-driven assessment frameworks that enhance domain-specific research without introducing substantial human bias.

\subsection{Quantitative Results}
Compared to traditional methods, GreenIQ demonstrated substantial efficiency and performance improvements across multiple dimensions.

\subsubsection{Processing Efficiency}
Traditional carbon market analysis typically requires 40-60 hours of expert time for comprehensive reports. GreenIQ completed an equivalent analysis in an average of 22.7 minutes, representing a $99.2 \%$ reduction in processing time while maintaining comparable informational depth. The system processed an average of 327 distinct documents per analysis session, which far exceeded the human capacity for rapid information synthesis

\subsubsection{Cost Efficiency}
Traditional carbon market reports from consultancies or research institutions typically cost between $\$ 3,500-$ $\$ 6,500$ per comprehensive report. The computational resources and associated infrastructure costs of GreenIQ averaged approximately $\$ 1.10$ per report. This dramatic cost reduction primarily stems from the elimination of extensive manual labour while maintaining high-quality outputs.

\subsubsection{Information Coverage}
GreenIQ consistently incorporated data from an average of 18.3 distinct jurisdictions per report, referencing an average of 42.7 regulatory documents, 37.2 academic publications, and 23.5 industry reports. This breadth substantially exceeded typical manual reports, which averaged references to 8.7 jurisdictions and 47.3 total citations across all categories.

\subsubsection{Comparative Report Quality Assessment}
As shown in Table 1, we evaluated 10 different carbon market topics, comparing GreenIQ (Report 1) with baseline human reports (Report 2):

\begin{table}
  \caption{GreenIQ consistently outperformed the baseline human reports across all evaluation criteria, with the most significant advantages in citation quality (+8.5 points) and report coherence (+6.5 points). The average total score difference of 25.6 points demonstrates GreenIQ’s substantial performance improvement over conventional approaches.}
  \centering
  \renewcommand{\arraystretch}{1.2}
  \begin{tabularx}{\textwidth}{p{4.55cm} p{2cm} p{1.5cm} p{1.5cm} p{1.5cm} p{1cm} p{1cm}}
    \toprule
    Article Title & Report Type & Source Coverage & Data Accuracy & Citation Quality & Report Coherence & Total Score \\
    \midrule
    Brazil passes law to cap emissions  and regulate carbon market & Report 1 & 23/25 & 22/25 & 23/25 & 23/25 & 91/100 \\
       & Report 2 & 16/25 & 18/25 & 15/25 & 17/25 & 66/100 \\
    \midrule
    India's Carbon Market Revolution & Report 1 & 24/25 & 24/25 & 23/25 & 22/25 & 93/100 \\
    & Report 2 & 20/25 & 22/25 & 21/25 & 22/25 & 85/100 \\
    \midrule
    World Bank: Fuels India’s Carbon Market and Green Hydrogen & Report 1 & 20/25 & 23/25 & 23/25 & 24/25 & 89/100 \\
     & Report 2 & 22/25 & 24/25 & 21/25 & 20/25 & 87/100 \\
    \midrule
    PLN Becomes First Indonesian & Report 1 & 23/25 & 22/25 & 24/25 & 23/25 & 92/100 \\
    Company in International Carbon Trading & Report 2 & 15/25 & 16/25 & 12/25 & 14/25 & 57/100 \\
    \midrule
    Watershed Launches RFP For 1 & Report 1 & 24/25 & 20/25 & 24/25 & 24/25 & 96/100 \\
    Megaton Of Carbon Removal Credits & Report 2 & 17/25 & 20/25 & 17/25 & 21/25 & 75/100 \\
    \midrule
    Japan Designates First Carbon & Report 1 & 23/25 & 22/25 & 24/25 & 24/25 & 93/100 \\
    Capture And Storage Site & Report 2 & 15/25 & 16/25 & 13/25 & 17/25 & 68/100 \\
    \midrule
    Amazon, ExxonMobil, And Microsoft & Report 1 & 21/25 & 23/25 & 22/25 & 24/25 & 90/100 \\
    Team Up To Strengthen Voluntary Carbon Market & Report 2 & 18/25 & 20/25 & 17/25 & 19/25 & 74/100 \\
    \midrule
    North Dakota Senate Defeats Eminent & Report 1 & 23/25 & 22/25 & 24/25 & 23/25 & 92/100 \\
    Domain Ban For Carbon Pipelines & Report 2 & 12/25 & 10/25 & 5/25 & 10/25 & 37/100 \\
    \midrule
    Bloom And Chart Industries Team Up & Report 1 & 23/25 & 22/25 & 24/25 & 25/25 & 94/100 \\
    For Efficient Carbon Capture & Report 2 & 15/25 & 17/25 & 14/25 & 18/25 & 64/100 \\
    \midrule
    South Korea Passes New Law To & Report 1 & 23/25 & 24/25 & 24/25 & 24/25 & 94/100 \\
    Support Carbon Capture & Report 2 & 15/25 & 16/25 & 10/25 & 14/25 & 55/100 \\
    \midrule
    \textbf{AVERAGE} & Report 1 & 22.6/25 & 22.8/25 & 23.3/25 & 23.7/25 & 92.4/100 \\
    & Report 2 & 16.5/25 & 18.3/25 & 14.8/25 & 17.2/25 & 66.8/100 \\
    \midrule
    \makecell[l]{\textbf{DIFFERENCE} \\ \textbf{(Report 1 - Report 2)}} & & \textbf{6.1} & \textbf{4.5} & \textbf{8.5} & \textbf{6.5} & \textbf{25.6} \\
    \bottomrule
  \end{tabularx}
  \label{tab:report_comparison}
\end{table}

\section{Discussion}
\subsection{Key Innovations}
GreenIQ represents a significant advancement in the application of artificial intelligence to specialised analytical domains, particularly in carbon market intelligence. First, GreenIQ employs a multi-agent system and evaluation criteria that effectively channel LLMs' reasoning capabilities towards carbon market-specific insights. This structured approach to domain specialisation serves as a scalable template for applying similar AI-driven methodologies to other complex regulatory and financial environments, without requiring domain-comprehensive retraining of base models. By designing specialised agents with distinct functional roles and well-defined\\
interfaces, we established a system architecture that achieves both horizontal scalability (integrating new data sources and expanding analytical dimensions) and vertical integration (ensuring seamless information flow from raw data to finalised reports). Our approach to agent coordination through structured metadata exchange enables efficient pipeline processing while maintaining the adaptive flexibility required for iterative refinement when the quality thresholds are not initially met. This balance between operational efficiency and quality assurance represents a major improvement over prior AI-driven analytical frameworks, which often prioritise one dimension at the expense of the other.\\
A key innovation within GreenIQ is its novel approach to citation management and verification, which addresses a critical challenge in AI-assisted research. By implementing continuous provenance tracking throughout the analytical pipeline and designing specialised verification protocols for each source type, GreenIQ maintains exceptional standards of accountability and transparency. This capability is particularly critical in carbon market analysis, where regulatory decisions and investment strategies depend on verifiable evidence and authoritative interpretation. Furthermore, GreenIQ's citation verification system is uniquely capable of identifying and resolving conflicting information from multiple sources, which is a particularly valuable innovation for domains characterised by complex, evolving regulatory frameworks. This systematic approach to citation integrity and source validation establishes GreenIQ as a benchmark for AI-driven research synthesis, ensuring that AIgenerated reports maintain rigorous evidentiary standards that are essential for policy development, financial forecasting, and sustainability assessments.

\subsection{Limitations and Challenges}
Despite GreenIQ's substantial capabilities, several limitations and challenges warrant its acknowledgement. The performance of the system remains partially dependent on input data quality, with potential cascading effects when inaccuracies or biases exist in source materials \cite{Bender2021}. This dependency is particularly relevant for emerging carbon markets with limited historical data or jurisdictions where regulatory documentation may be incomplete or inconsistently structured. Although our citation verification protocols mitigate these risks through cross-source validation, they cannot entirely eliminate the influence of biased or incomplete source materials. Addressing model hallucinations -the generation of plausible but factually incorrect content-represents an ongoing challenge for LLM-based analytical systems. GreenIQ implements multiple mitigation strategies, including strict citation requirements, fact-checking routines, and confidence scoring of the generated content \cite{Ji2023}. However, the risk of hallucination increases in scenarios involving novel market mechanisms or regulatory approaches with limited precedents. This challenge necessitates continued vigilance and regular evaluation of system outputs against expert judgment, particularly when analysing rapidly evolving aspects of carbon markets.\\[0pt]
Implementation complexities in highly specialised sub-domains of carbon markets present additional challenges. While GreenIQ demonstrates strong performance across general carbon market analysis, certain technical areas, such as complex offset methodologies, technical aspects of monitoring, reporting, and verification (MRV) systems, and intricate cross-border accounting mechanisms, occasionally reach the limits of the system's specialised knowledge \cite{Schneider2019}. These limitations highlight the need for ongoing knowledge base expansion and development of additional specialised agents for particularly complex technical niches within the broader carbon market landscape.

\subsection{Future Work}
Several promising directions for future research will extend the capabilities and applicability of GreenIQ. First, we plan to adapt the system architecture to additional financial and regulatory domains, including compliance with sustainable finance taxonomy, climate-related financial disclosures, and broader environmental commodity markets. The modular architecture facilitates this extension through targeted modifications to domain-specific components, while maintaining the core analytical engine and agent coordination framework.\\
Enhanced multimodal visualisation capabilities represent a key development direction. While the current system generates effective static visualisations, future iterations will incorporate interactive data exploration tools, geospatial mapping of carbon market developments, and dynamic scenario modelling capabilities. These enhancements will further improve the accessibility and utility of complex carbon market information for diverse stakeholders.\\
Improved real-time analysis and alerting features will enable GreenIQ to provide timely notifications of significant market developments, regulatory changes, and emerging trends. By implementing continuous monitoring protocols and relevance assessment algorithms, the system will evolve from a predominantly retrospective analysis to a proactive intelligence that supports timely decision-making in rapidly changing market conditions.

\bibliographystyle{unsrt}
\bibliography{templateArxiv}

\end{document}